\documentclass[runningheads]{llncs}

\usepackage[T1]{fontenc}
\usepackage[utf8]{inputenc}
\usepackage{graphicx}
\usepackage{amsmath}
\usepackage{amssymb}
\usepackage{amsfonts}
\usepackage{booktabs}
\usepackage{float}
\usepackage{multirow}
\usepackage{hyperref}
\usepackage{url}
\usepackage{xcolor}
\usepackage{microtype}

\spnewtheorem{observation}{Observation}{\bfseries}{\itshape}
\spnewtheorem{finding}{Empirical Finding}{\bfseries}{\itshape}
\spnewtheorem{consequence}{Consequence}{\bfseries}{\itshape}

\newcommand{\sstar}{s^{*}}
\newcommand{\R}{\mathbb{R}}
\newcommand{\E}{\mathbb{E}}
\newcommand{\Var}{\mathrm{Var}}

\begin{document}

\title{Confidence Scores in Open-Vocabulary Detection\\
Are a Biased Mixture of Scale and Semantics}

\titlerunning{Confidence Score Bias in Open-Vocabulary Detection}

\author{Yi Tang Soon\inst{1}\orcidID{0009-0009-8889-8989} \and Jun-Wei Hsieh\inst{2}\orcidID{0000-0002-5477-4891}}

\authorrunning{Y.T. Soon and J.W. Hsieh}

\institute{
  Institute of Intelligence Systems,
  National Yang Ming Chiao Tung University, Tainan, Taiwan\\
  \email{esthersoon2002.ai14@nycu.edu.tw}
  \and
  College of Artificial Intelligence and Green Energy,
  National Yang Ming Chiao Tung University, Hsinchu, Taiwan\\
  \email{jwhsieh@nctu.edu.tw}
}
\maketitle

\let\thefootnote\relax\footnotetext{This version of the article has been accepted for publication, after peer review but is not the Version of Record and does not reflect post-acceptance improvements, or any corrections. The Version of Record will be available online at Springer Link.}

\begin{abstract}
Foundation models such as CLIP~\cite{radford2021clip} have enabled
open-vocabulary object detectors that generalise to novel categories
via vision-language similarity.
However, the confidence scores these detectors produce are not
reliable localization probability estimates: they conflate visual
scale and semantic query specificity with the true detection signal.
Through controlled experiments on COCO across three
foundation-model-based detectors (GroundingDINO, OWL-ViT,
YOLO-World), with the scale-bias finding further replicated on
LVIS (1{,}203 categories) using GroundingDINO, we show that
$s = \cos(v,t)$ is a biased mixture of two effects.
Scale bias ($\hat{\alpha} = {+}0.064$, $r = 0.579$,
$p = 1.29 \times 10^{-58}$) systematically inflates scores for
large objects.
Semantic bias ($\hat{\beta} = {-}0.705$,
$p = 5.23 \times 10^{-41}$) suppresses scores for generic queries.
Both biases are structurally inevitable from CLIP's image-level
pretraining. Threshold adjustment cannot remove them: oracle
per-scale thresholding yields $\Delta\text{F1} = {+}0.001$ for
small objects versus ${+}0.102$ for large.
A parameter-free temperature scaling correction improves
small-object Recall@10 by 19.6\% ($p < 0.01$) without retraining.
This comes at a modest, measurable cost to pooled-ranking
precision, so the bias is partially, not freely, reversible at
inference time.
These findings reveal a fundamental limitation of adapting
image-level foundation models to region-level detection tasks.

\end{abstract}

\section{Introduction}
\label{sec:intro}

Foundation models such as CLIP~\cite{radford2021clip},
ALIGN~\cite{jia2021align}, and SigLIP~\cite{zhai2023siglip} have
transformed computer vision by providing powerful image-text
representations that generalise across tasks.
Open-vocabulary object detectors build on these models to localise
objects described by arbitrary text queries.
GroundingDINO~\cite{liu2023grounding},
OWL-ViT~\cite{minderer2022simple}, and
YOLO-World~\cite{cheng2024yolo} all achieve strong zero-shot
performance this way.
The standard output of these systems is a scalar confidence score
$s \in [0,1]$ per detection, derived from cosine similarity in
CLIP space.
Practitioners use this score to filter detections by a fixed
threshold.
The implicit assumption is that $s$ measures localization
certainty: a detection with $s = 0.7$ should be correct 70\% of
the time, regardless of object size or query specificity.
We show this assumption is fundamentally false.

Consider applying such a detector to a scene.
At a threshold of 0.3, small objects are systematically discarded.
This is not because the detector failed to localise them: it
produced correct bounding boxes, but with confidence scores of
0.08, 0.11, and 0.09.
Lowering the threshold to rescue small objects floods the output
with large-object false positives.
The root cause is not a model failure but a structural property
of $s = \cos(v,t)$ that conflates two independent signals.

We identify and quantify two systematic biases:

\noindent\textbf{Scale bias} ($\hat{\alpha} = {+}0.064$,
$r = 0.579$, $p = 1.29 \times 10^{-58}$).
Large objects receive systematically higher confidence scores than
small objects for identical queries.
Averaging CLIP features over fewer pixels yields a noisier, less
concentrated region direction, which pulls the normalised cosine
similarity away from its noise-free value.

\noindent\textbf{Semantic bias} ($\hat{\beta} = {-}0.705$,
$p = 5.23 \times 10^{-41}$).
For fixed visual content, generic queries (``an object'') produce
lower scores than specific queries (``a dog''), because specific
queries occupy tighter regions in CLIP embedding space.

Scale bias is replicated across three detectors and on LVIS.
Semantic bias is demonstrated on GroundingDINO across 8 categories.
Both derive theoretically from CLIP's image-level contrastive
pretraining objective. Our contributions are:
\begin{enumerate}
\item A formal additive decomposition
  $s \approx \alpha\!\cdot\!\phi(a) + \beta\!\cdot\!\psi(d_\mathrm{sem}) + \varepsilon$
  with identified coefficients, showing that foundation model
  confidence scores are structurally biased for region-level tasks.
\item Controlled experiments confirming scale bias across three
  detectors ($r \in [0.36, 0.64]$) and semantic bias on
  GroundingDINO across 8 categories (paired $t > 16$).
\item A $2\times2$ factorial experiment revealing a floor effect
  that drowns out the localization signal for small objects.
\item A parameter-free test-time correction that improves small
  object Recall@10 by 20\% without retraining, with its
  precision and calibration costs measured rather than assumed.
\end{enumerate}

\section{Related Work}
\label{sec:related}

\subsubsection{Foundation Models for Detection.}
Open-vocabulary detection builds on vision-language foundation
models.
Early work~\cite{zareian2021ovrcnn} used captions for novel
categories.
Subsequent methods include ViLD~\cite{gu2021vild},
GLIP~\cite{li2022glip}, RegionCLIP~\cite{zhou2022regionclip},
OWL-ViT/v2~\cite{minderer2022simple,minderer2023scaling},
GroundingDINO~\cite{liu2023grounding},
YOLO-World~\cite{cheng2024yolo}, and
Bangalath et al.~\cite{bangalath2022bridging}.
Concurrently, \cite{zheng2024aggregation} proposes a training-free
confidence aggregation scheme that boosts open-vocabulary detection
accuracy by combining scores across proposals.
All of these share the same confidence interface: cosine similarity
in CLIP space.
Our work questions whether this scalar is a meaningful quantity for
downstream applications.
It also provides a structural account of why such
aggregation-style fixes are needed in the first place.

\subsubsection{Limitations of Image-Level Pretraining.}
CLIP's image-level contrastive objective has known
spatial limitations.
Prior work shows failures on compositional
reasoning~\cite{yuksekgonul2023winoground},
spatial understanding~\cite{zhang2024beyond,parcalabescu2021seeing},
and fine-grained visual tasks~\cite{tong2024eyes}.
DeCLIP~\cite{li2022declip} and FLAVA~\cite{singh2022flava} explore
alternatives but retain image-level alignment.
We show that the confidence score aggregation mechanism introduces
additional systematic biases independent of feature quality.

\subsubsection{Confidence Calibration.}
Calibration is studied for
classification~\cite{guo2017calibration,nixon2019measuring},
dataset shift~\cite{ovadia2019can}, and
detection~\cite{kuppers2020multivariate}.
Temperature scaling~\cite{guo2017calibration} and label
smoothing~\cite{szegedy2016rethinking} are post-hoc logit fixes
that cannot address structural entanglement.
No prior work identifies the structural sources of miscalibration
in foundation-model-based open-vocabulary detectors.

\subsubsection{Small Object Detection.}
Small object detection is addressed through feature
pyramids~\cite{lin2017feature}, single-stage
detectors~\cite{liu2016ssd,redmon2016yolo}, and specialised
architectures~\cite{chen2021you,ren2015faster,carion2020detr}.
We provide a new structural explanation for why
foundation-model-based open-vocabulary detectors systematically
underperform on small objects.

\section{Formal Formulation}
\label{sec:formulation}

Let $v \in \R^d$ denote the visual region feature and
$t \in \R^d$ the text embedding of query $q$.
Foundation-model-based open-vocabulary detectors compute:
\begin{equation}
  s = \cos(v, t) = \frac{v \cdot t}{\|v\|\,\|t\|}.
  \label{eq:score}
\end{equation}
In practice, each detector implements Eq.~\eqref{eq:score} through
a learned, sigmoid-activated projection head rather than an
unprocessed cosine value.
We use $\cos(v,t)$ as the shared formal abstraction of this
interface throughout, since all three detectors reduce to a
monotone function of region-text similarity.
Throughout, $a$ denotes the candidate region's ground-truth pixel
area, the geometric proxy for object scale.
$d_\mathrm{sem}$ denotes the semantic specificity of query $q$.
We operationalise it as the CLIP text-embedding cosine distance
$d_\mathrm{sem} = 1 - \cos(t_q, t_{q_0})$ between $q$'s text
embedding and the embedding of the category's most specific query
$q_0$ (e.g.\ \texttt{"a dog"}).
By this definition, $d_\mathrm{sem}=0$ for the most specific
wording and increases for progressively more generic rewordings of
the same category (Sec.~\ref{sec:experiments}, Experiment~2).

\begin{definition}[Ideal Confidence Score]
The ideal confidence score is:
$\sstar = P(\mathrm{IoU}(\hat{b}, b^*) \geq 0.5 \mid v, t)$,
where $\hat{b}$ is the predicted box and $b^*$ the ground truth.
A score is well-calibrated if $s \approx \sstar$.
\end{definition}

We present two observations showing $s$ is not well-calibrated.

\begin{observation}[Scale Bias]
With category and query fixed, $\E[s \mid a_\mathrm{small}]
\ll \E[s \mid a_\mathrm{large}]$.
Across three detectors, Pearson $r \in [0.36, 0.64]$ between
$\log a$ and $s$ ($p < 10^{-20}$).
\end{observation}

\begin{observation}[Semantic Bias]
With visual content fixed,
$\E[s \mid t_\mathrm{specific}] \gg \E[s \mid t_\mathrm{generic}]$.
Paired $t = 16.5$, $p = 9.83 \times 10^{-49}$ across 8 categories.
\end{observation}

\begin{finding}[Additive Decomposition]
Let $\phi(a) = \log(a+1)$ be the log-area scale term, and let
$\psi$ be the identity map, $\psi(d_\mathrm{sem}) = d_\mathrm{sem}$,
so the semantic specificity level enters linearly.
The observed score decomposes as:
\begin{equation}
  \boxed{s \approx \alpha \cdot \phi(a)
         + \beta \cdot \psi(d_\mathrm{sem}) + \varepsilon} ,
  \label{eq:decomp}
\end{equation}
with $\hat{\alpha} = {+}0.064$ (from Exp.~1) and
$\hat{\beta} = {-}0.705$ (from Exp.~2), both
$p < 10^{-40}$.
The residual $\varepsilon$ contains the localization signal
$\sstar$ and noise.
Equation~\eqref{eq:decomp} is a first-order (leading-term)
approximation that linearises the multiplicative mechanism derived
in Sec.~\ref{sec:theory} (Eq.~\eqref{eq:scale_derive}).
It is intended to identify and quantify the two bias coefficients,
not to claim that $s$ is exactly additive in $\phi(a)$ and
$\psi(d_\mathrm{sem})$ everywhere.
The non-parallel interaction observed in Experiment~3 is consistent
with, rather than contradicting, this multiplicative origin.
As $a$ shrinks, the scale term drives $s$ toward zero and
compresses the room available for the semantic term to act.
This produces the floor effect reported in
Sec.~\ref{sec:experiments}.
\end{finding}

\begin{consequence}[Loss of Discriminative Power]
For small objects,
$\Var(\varepsilon \mid a_\mathrm{small}) \approx
\Var(s \mid a_\mathrm{small}) -
\hat{\alpha}^2 \Var(\phi(a_\mathrm{small}))
= 0.0138 - 0.0054 = 0.0084$,
insufficient to separate true from false positives
(confirmed empirically: oracle thresholding yields
$\Delta\mathrm{F1} = {+}0.001$ for small vs.\ ${+}0.102$ for large).
\end{consequence}

\subsection{Why Entanglement Is Structurally Inevitable}
\label{sec:theory}

\noindent\textbf{Scale bias: directional concentration.}
Since $s=\cos(v,t)$ is invariant to the norm of $v$ by
construction, scale bias cannot arise from $\|v\|$ itself.
It must arise from how scale affects the direction of $v$.
Write the unnormalised region feature as
$v_\mathrm{raw} = \mu + \bar\eta(R)$, where $\mu$ is the
noise-free, category-aligned direction and
$\bar\eta(R) = \frac{1}{|R|}\sum_{p\in R}\eta(p)$ is the average of
zero-mean, variance-$\sigma_f^2$ pixel-level feature noise over
region $R$.
By the law of large numbers, $\Var(\bar\eta(R)) \propto
\sigma_f^2/|R|$, so smaller regions yield a noisier estimate of
$\mu$.
After normalisation, the resulting direction
$v = v_\mathrm{raw}/\|v_\mathrm{raw}\|$ deviates from
$\mu/\|\mu\|$ by a small, zero-mean angle $\delta$, with
$\Var(\delta) \propto \sigma_f^2/(|R|\,\|\mu\|^2)$.
Writing $\theta = \angle(\mu, t)$ for the noise-free
category-query alignment (the localisation signal $\sstar$ is
monotone in $\cos\theta$) and expanding $\cos(\theta+\delta)$ to
second order in $\delta$:
\begin{equation}
  \E[s \mid a] \;\approx\; \cos\theta\left(1 - \tfrac12\Var(\delta)\right)
  \;\approx\; \cos\theta\left(1 - \frac{\kappa}{2a}\right),
  \label{eq:scale_derive}
\end{equation}
where $a = |R|$ is the region's pixel area and
$\kappa \propto \sigma_f^2/\|\mu\|^2$ is a constant governing the
local feature-noise level, treated as approximately fixed for a
given category-query pair. For $\cos\theta > 0$ (the relevant
regime for true matches), Eq.~\eqref{eq:scale_derive} is
monotonically increasing in $a$ and asymptotes to the noise-free
alignment $\cos\theta$ as $a \to \infty$. This is exactly the angular-concentration behaviour of a von
Mises--Fisher-distributed direction estimate, whose concentration
parameter grows with the number of pixels averaged.
Scale bias is therefore a structural consequence of spatial
pooling acting on the direction of the region embedding.

\noindent\textbf{Semantic bias: CLIP alignment geometry.}
CLIP's contrastive loss trains specific query embeddings (e.g.,
``a dog'') to align tightly with concept features.
Generic queries (``an object'') instead align diffusely with all
objects.
For any fixed $v$:
$\cos(v, t_{q_0}) > \cos(v, t_{q_k})$ for $k > 0$,
predicting the confidence drop confirmed in Observation~2.

Both biases arise from the same root: $s = \cos(v,t)$
optimises image-level semantic similarity, not region-level
localization certainty.

\section{Experiments}
\label{sec:experiments}

\subsubsection{Setup.}
We evaluate GroundingDINO-SwinT~\cite{liu2023grounding},
OWL-ViT-B/32~\cite{minderer2022simple}, and
YOLO-World-S~\cite{cheng2024yolo} off-the-shelf on COCO 2014
val~\cite{lin2014coco}, 8 categories, 80 images per category
(640 pairs total).
Scale groups: small ($a < 32^2$), medium ($32^2$--$96^2$),
large ($a \geq 96^2$).
Representative instances at each scale include, e.g., a distant bird (small), a chair viewed across a room (medium), and a close-range car or
truck filling much of the frame (large).
Unless otherwise noted, inference uses box/text threshold 0.01,
which retains the weak detections needed for the
precision--recall and oracle-thresholding analyses of
Sec.~\ref{sec:consequences}.
Experiment~1's GroundingDINO correlation (Table~\ref{tab:cross},
Fig.~\ref{fig:scale}) instead uses threshold 0.05/0.10, which is
sufficient for the sanity-check correlation it reports.

\subsubsection{Experiment 1: Scale Bias.}
\begin{figure}[H]
  \centering
  \includegraphics[width=\textwidth]{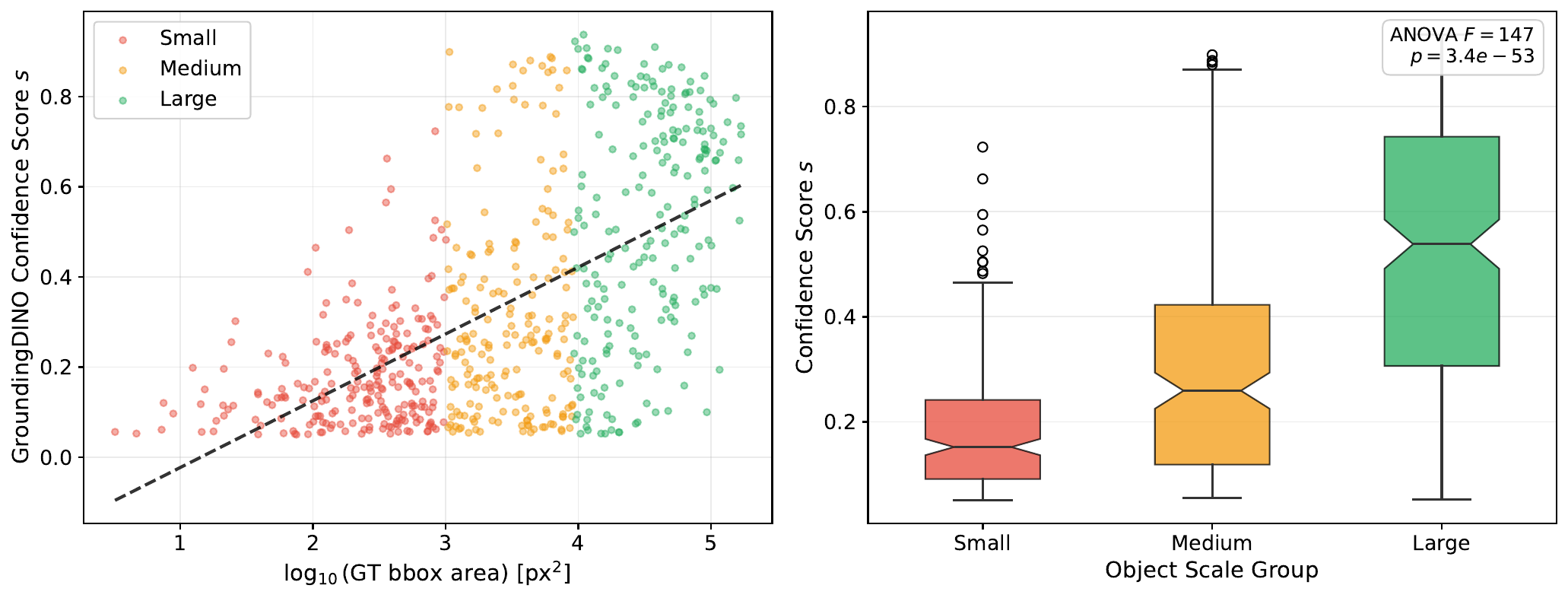}
  \caption{Scale bias in GroundingDINO (COCO, 8 categories).
    \textit{Left}: Confidence vs.\ $\log(\text{area})$;
    regression line $r=0.579$, $p=1.29\times10^{-58}$.
    \textit{Right}: Confidence distributions by scale group;
    ANOVA $F=146.92$, $p=3.40\times10^{-53}$.
    Text query held constant throughout.}
  \label{fig:scale}
  \vspace{-5mm}
\end{figure}

With text query fixed to \texttt{"a [category]"}, mean confidence
rises monotonically: 0.180 (small), 0.317 (medium), 0.520 (large)
(ANOVA $F=146.92$, $p=3.40\times10^{-53}$; Fig.~\ref{fig:scale}).
Per-category $r \in [0.296, 0.666]$, all $p < 0.01$.
Table~\ref{tab:cross} shows the bias is universal across all three
detectors.\footnote{Even the smallest $p$-value ($4.56\times10^{-72}$ for
YOLO-World) is far from machine epsilon
($\approx 1.1\times10^{-16}$). It comes from the analytic
Student-$t$ CDF, not floating-point subtraction, and reflects a
strong correlation ($|r|>0.6$) over hundreds of samples.}

\begin{table}[H]
  \vspace{-3mm}
\caption{Scale-confidence Pearson $r$ across three detectors.
All values positive and significant ($p < 10^{-20}$).}
\label{tab:cross}
\centering
\begin{tabular}{lcccc}
\toprule
Detector & $r$ & $p$ & Small mean & Large mean \\
\midrule
GroundingDINO & 0.579 & $1.29{\times}10^{-58}$ & 0.180 & 0.520 \\
OWL-ViT       & 0.363 & $3.18{\times}10^{-21}$ & 0.085 & 0.279 \\
YOLO-World    & 0.637 & $4.56{\times}10^{-72}$ & 0.145 & 0.615 \\
\bottomrule
\end{tabular}
  \vspace{4mm}
\end{table}
\newpage

\begin{figure}[H]
  \centering
  \includegraphics[width=\textwidth]{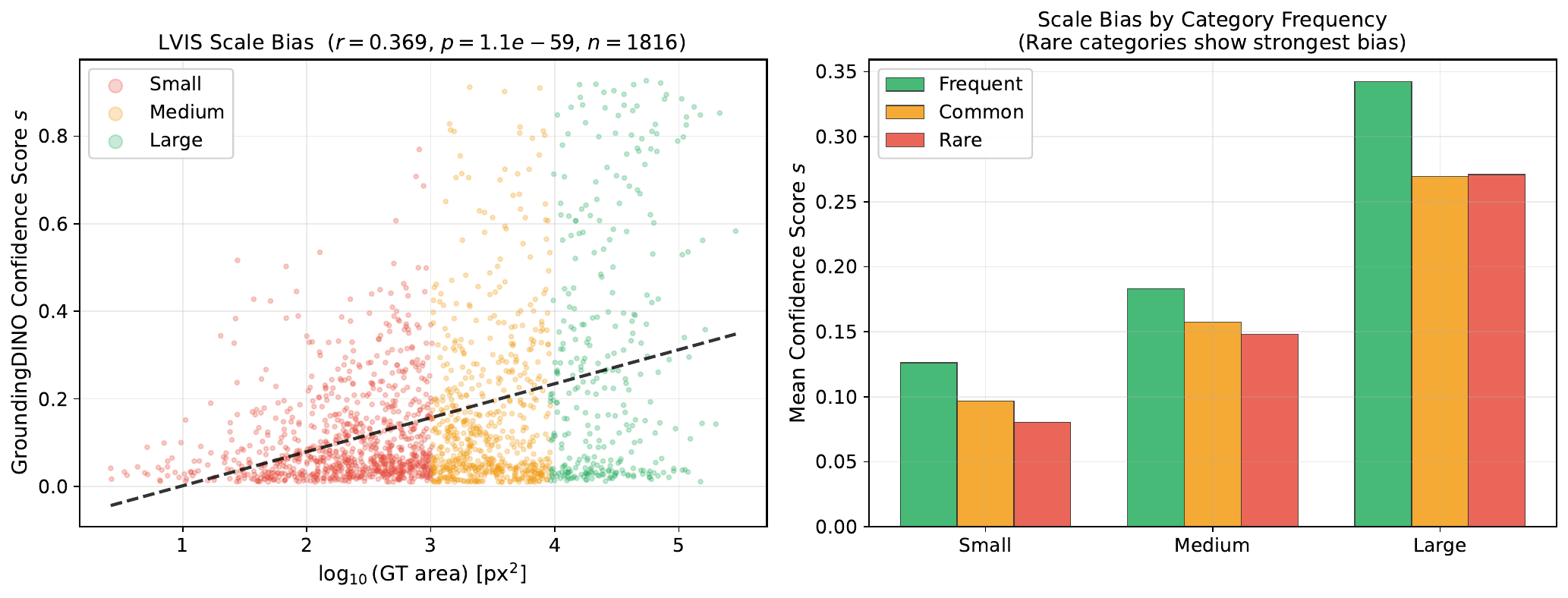}
  \caption{Scale bias replication on LVIS v1 val (1{,}203 categories,
    $n=1{,}816$).
    \textit{Left}: Confidence vs.\ $\log(\text{area})$;
    $r=0.369$, $p=1.08\times10^{-59}$.
    \textit{Right}: Mean confidence by scale group and category
    frequency. Rare categories show the strongest bias ($r=0.491$),
    suggesting that objects underrepresented in CLIP pretraining
    data are more susceptible to scale-induced score distortion.}
  \label{fig:lvis}
\end{figure}

\paragraph{Replication on LVIS, Fig.~\ref{fig:lvis}}
To test whether scale bias generalises beyond 8 hand-picked
categories, we replicate Experiment~1 on LVIS v1
val~\cite{gupta2019lvis}, which covers 1{,}203 categories across
19{,}809 images.
Using GroundingDINO on 1{,}816 sampled annotations,
scale bias replicates strongly: $r = 0.369$,
$p = 1.08 \times 10^{-59}$, ANOVA $F = 135.90$,
$p = 1.01 \times 10^{-55}$.
Mean confidence follows the same direction as COCO:
0.110 (small), 0.167 (medium), 0.301 (large).
Notably, scale bias is stronger for rare categories
($r = 0.491$) than for frequent ($r = 0.386$) or common
($r = 0.351$) categories.
This suggests that objects with fewer training examples in CLIP's
pretraining corpus are more susceptible to scale-induced confidence
distortion.
This finding is particularly relevant for foundation model
deployment in long-tail recognition scenarios.

\subsubsection{Experiment 2: Semantic Bias.}
\begin{figure}[ht]
  \centering
  \includegraphics[width=\textwidth]{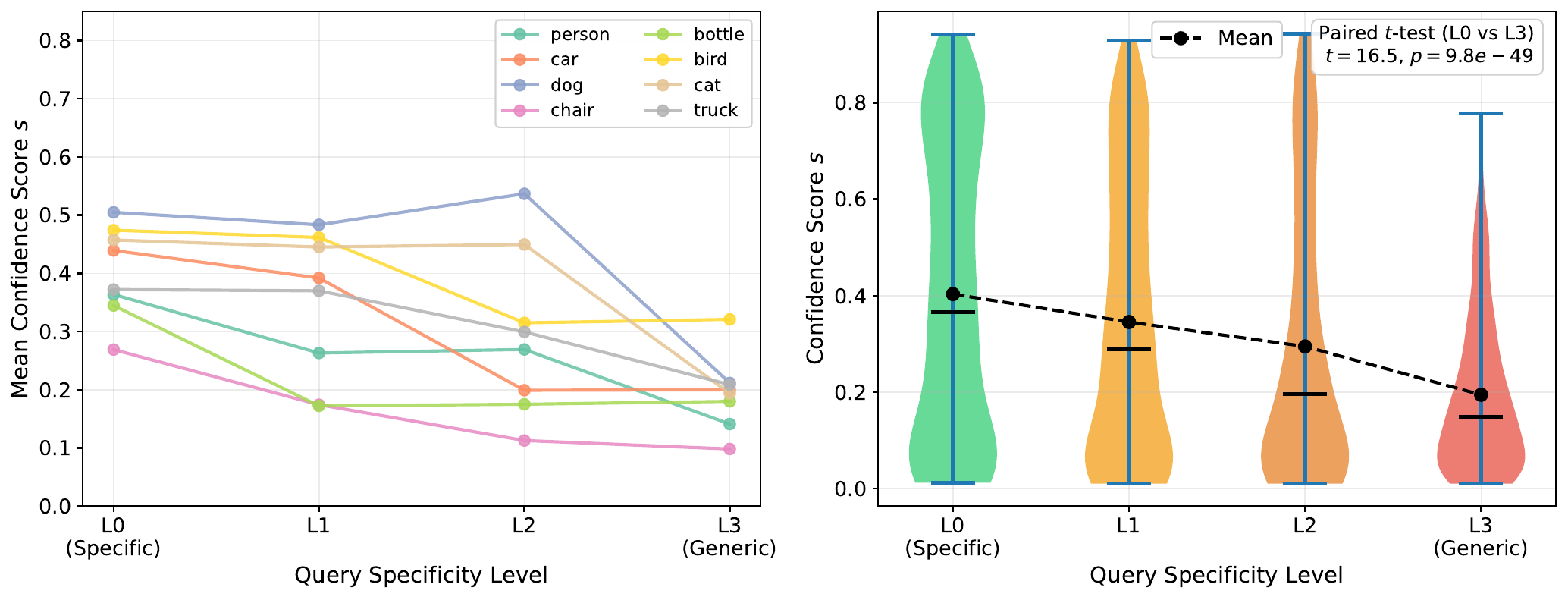}
  \caption{Semantic bias.
    \textit{Left}: Per-category mean confidence across query levels.
    \textit{Right}: Pooled violin plots.
    Paired $t=16.5$, $p=9.83\times10^{-49}$; visual content fixed.}
  \label{fig:semantic}
  \vspace{-2.5em}
\end{figure}

With visual content fixed (same image, same box, large objects
only), queries vary from specific ($q_0$) to generic ($q_3$) along
a fixed four-level hierarchy designed to be increasingly generic.
For category \texttt{dog}, this hierarchy is
L0\,=\,\texttt{"a dog"} (category name),
L1\,=\,\texttt{"a pet"} (functional category),
L2\,=\,\texttt{"an animal"} (superordinate class), and
L3\,=\,\texttt{"an object"} (maximally generic).
The same category-to-functional-to-superordinate-to-generic pattern
is applied to all 8 categories. The semantic-distance term
$d_\mathrm{sem}$ entering the decomposition (Eq.~\eqref{eq:decomp})
is the CLIP cosine distance from each level's query to the
category's L0 query (Sec.~\ref{sec:formulation}), not the ordinal
level itself.
Confidence drops across all 8 categories:
L0: 0.403 $\to$ L3: 0.194 (paired $t=16.5$,
$p=9.83\times10^{-49}$; Fig.~\ref{fig:semantic}).
CLIP distances are non-monotone for 3 of 8 categories,
reflecting corpus statistics rather than taxonomic hierarchy.

\subsubsection{Experiment 3: Two-Signal Interaction.}
\begin{figure}[ht]
  \centering
  \includegraphics[width=\textwidth]{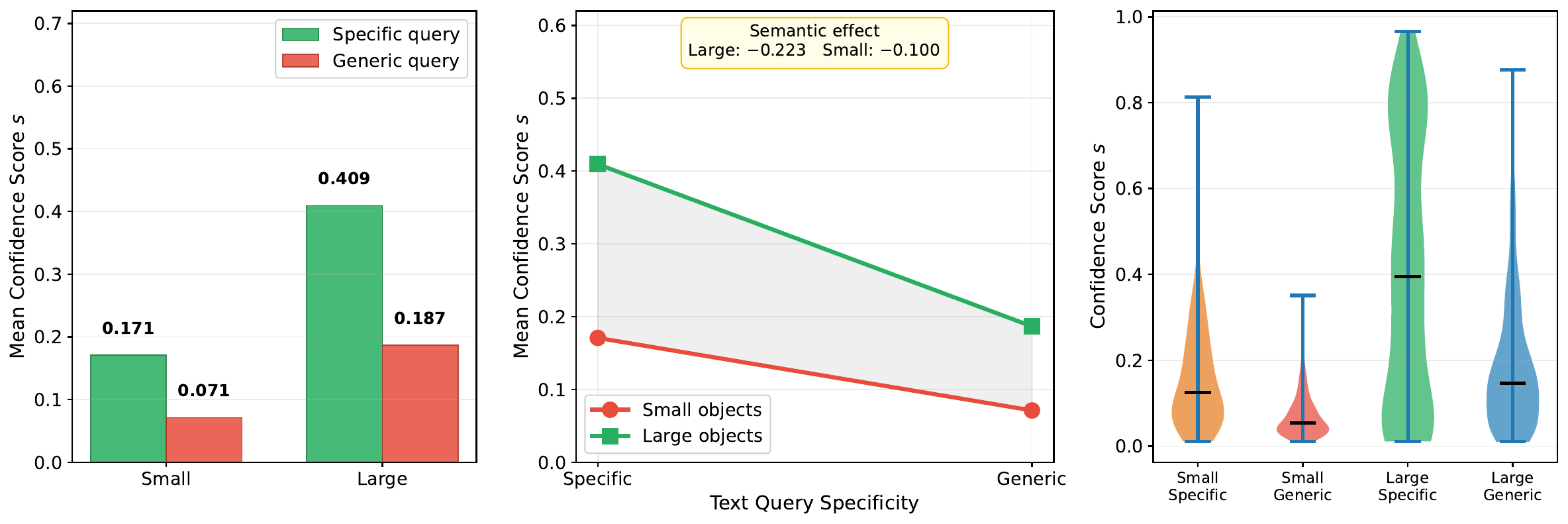}
  \caption{$2\times2$ interaction.
    \textit{Left}: Cell means.
    \textit{Center}: Interaction plot; non-parallel lines reveal a
    floor effect for small objects.
    \textit{Right}: Full distributions per cell.}
  \label{fig:2x2}
\end{figure}

A $2\times2$ factorial design (scale $\times$ specificity)
confirms both main effects ($p < 10^{-58}$; Fig.~\ref{fig:2x2},
Table~\ref{tab:2x2}).
The semantic effect is $2.2\times$ larger for large objects
($\Delta = {+}0.223$) than small ($\Delta = {+}0.100$).
Scale bias saturates small-object scores near zero, suppressing the semantic signal.

\begin{table}[ht]
\caption{$2\times2$ cell means: mean confidence score $s$
(GroundingDINO), averaged over the same 8 COCO categories used in
Experiments~1--2, for each combination of object-scale group
(large, $a\geq96^2$; small, $a<32^2$) and query specificity
(specific query L0 vs.\ generic query L3, defined in
Experiment~2). Both main effects significant ($p < 10^{-58}$).}
\label{tab:2x2}
\centering
\begin{tabular}{lcc}
\toprule
               & Specific query & Generic query \\
\midrule
Large objects  & 0.409 & 0.187 \\
Small objects  & 0.171 & 0.071 \\
\bottomrule
\end{tabular}
\end{table}

\section{Practical Consequences}
\label{sec:consequences}

\subsubsection{Precision-Recall and Threshold Gap.}
\begin{figure}[h]
  \centering
  \includegraphics[width=\textwidth]{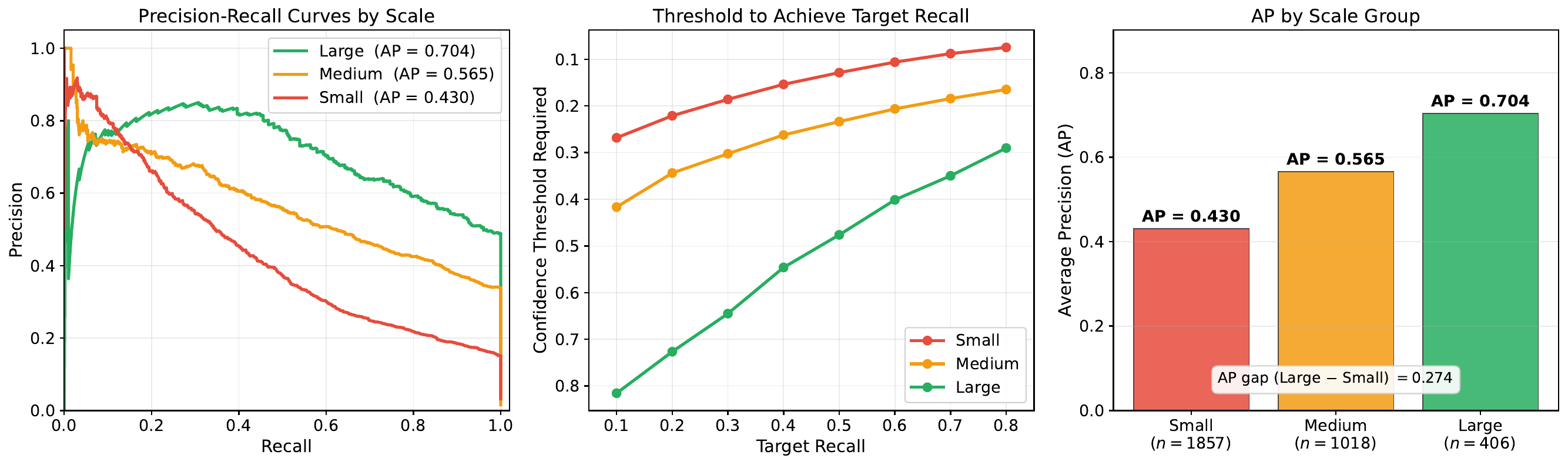}
  \caption{Practical cost of entanglement.
    \textit{Left}: PR curves (AP gap = 0.27).
    \textit{Center}: Threshold for target recall per scale.
    \textit{Right}: AP by scale group.}
  \label{fig:pr}
\end{figure}

We summarise detection quality per scale group with Average
Precision (AP): the area under the precision--recall curve,
equivalently the mean precision averaged over recall levels from
0 to 1.
AP is the standard scalar summary of a precision--recall curve.\footnote{Per-scale AP assigns each detection to a scale
bucket using its own predicted box area, and matches it only
against ground truth in that same bucket. A detection with no
same-scale ground truth in its image therefore counts as a false
positive.
This isolates each scale's precision--recall behaviour, but differs
from the COCO convention of matching against all ground truth in an
image and filtering by ground-truth scale post hoc.}
AP gaps and the confidence threshold $\tau$ required to reach a
target recall $R$, per scale group, are shown in
Table~\ref{tab:ap} and Fig.~\ref{fig:pr}.
The AP gap (large $-$ small) is 0.27.
Achieving 50\% recall for small objects requires a threshold
$3.7\times$ lower than for large objects (0.129 vs.\ 0.476).
Oracle per-scale thresholding provides no benefit for small
objects ($\Delta\mathrm{F1} = {+}0.001$) but substantial benefit
for large objects ($\Delta\mathrm{F1} = {+}0.102$).
This confirms the localization signal is unrecoverable by
threshold selection alone.

\begin{table}[h]
\caption{AP and the confidence threshold $\tau$ required to reach
target recall $R$ (e.g.\ $\tau(R{=}0.3)$ is the score cutoff
needed for 30\% recall), by scale group.}
\label{tab:ap}
\centering
\begin{tabular}{lcccc}
\toprule
Scale & AP & $\tau$ (R=0.3) & $\tau$ (R=0.5) & $\tau$ (R=0.7) \\
\midrule
Small  & 0.430 & 0.186 & 0.129 & 0.088 \\
Medium & 0.566 & 0.302 & 0.234 & 0.184 \\
Large  & 0.704 & 0.645 & 0.476 & 0.350 \\
\bottomrule
\end{tabular}
\end{table}

\newpage

\subsubsection{Calibration Analysis.}
\begin{figure}[h]
  \centering
  \includegraphics[width=\textwidth]{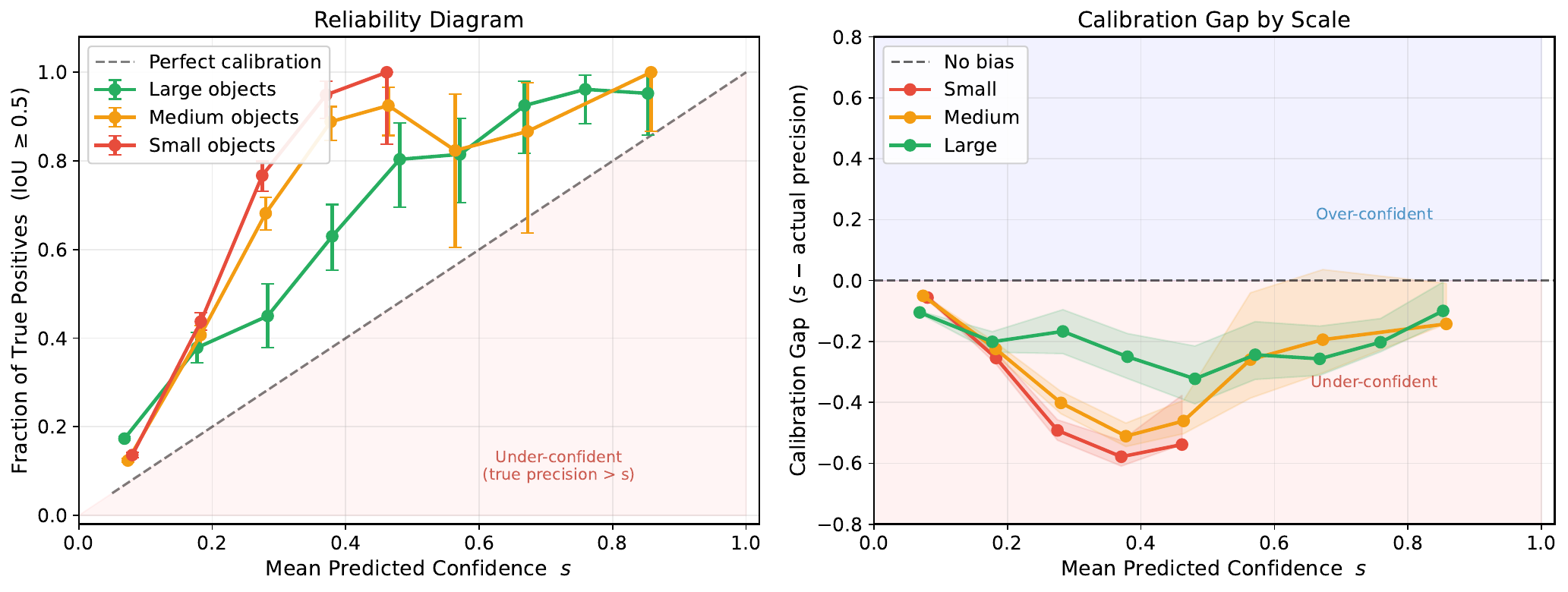}
  \caption{Scale-dependent miscalibration.
    \textit{Left}: Reliability diagram (90\% CI); small objects
    are furthest from the diagonal.
    \textit{Right}: Calibration gap; small objects show a larger
    negative gap (mean $-0.384$) than large (mean $-0.205$).}
  \label{fig:calibration}
\end{figure}

Reliability diagrams (Fig.~\ref{fig:calibration}), computed across
56,630 detections ($s \geq 0.05$), show all groups are
under-confident.
The gap is scale-dependent: $-0.384$ for small objects versus
$-0.205$ for large.
This confirms that scale bias suppresses scores below their true
localization probability.

\subsubsection{Parameter-Free Test-Time Correction.}
Scale bias creates a ranking unfairness in images with
mixed-scale objects: large objects dominate the top-$k$
detections regardless of localization quality.
We propose variance-matched temperature scaling to correct this:
\begin{equation}
  s_\mathrm{corr} = s^{1/T_g}, \quad
  T_g = \sqrt{\frac{\Var(\log s_g)}{\Var(\log s_\mathrm{large})}},
  \label{eq:correction}
\end{equation}
estimated without ground-truth labels ($T_\mathrm{small}=1.28$,
$T_\mathrm{medium}=1.06$, $T_\mathrm{large}=1.00$).

\begin{figure}[h]
\vspace{-10mm}
  \centering
  \includegraphics[width=0.7\textwidth]{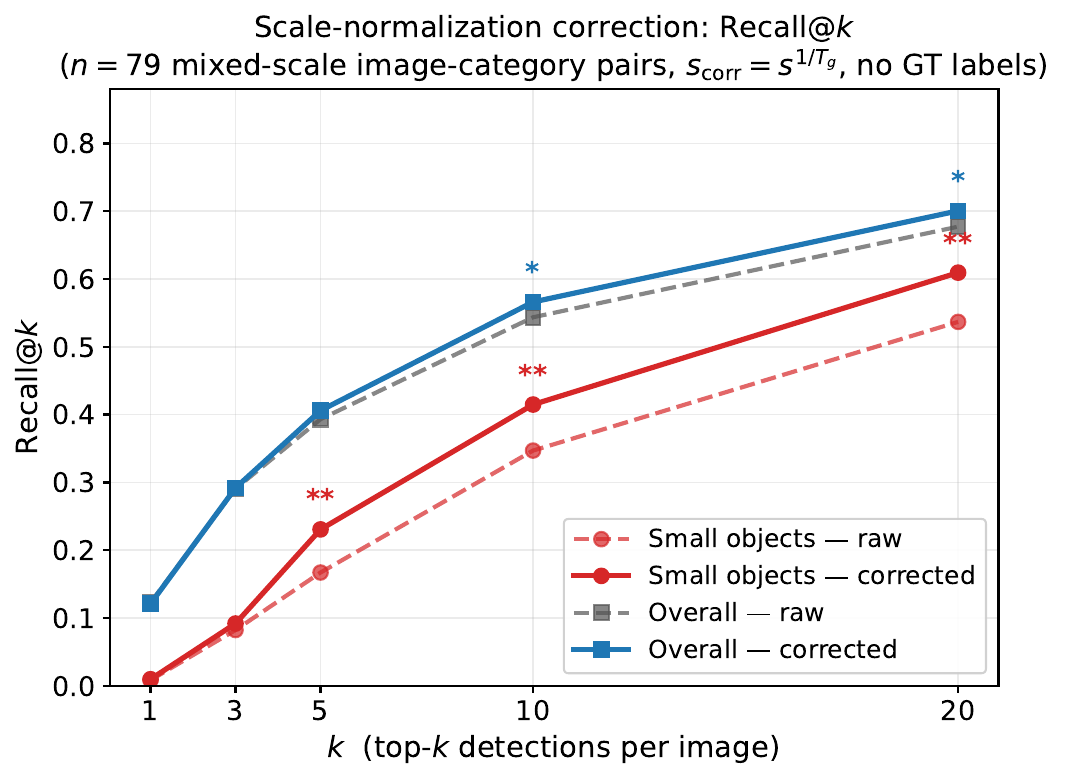}
  \caption{Scale-normalization correction on $n=79$ mixed-scale
    (image, category) pairs (images containing both small and
    large ground-truth objects). Solid: corrected; dashed: raw.
    Small-object Recall@$k$ improves significantly at all $k$
    ($^{**}p<0.01$); overall Recall@10 also improves
    ($^{*}p<0.05$). Large-object recall decreases modestly and
    non-significantly ($-6\%$, $p=0.10$).
    No ground-truth labels required.}
  \label{fig:correction}
\end{figure}

On $n=79$ mixed-scale pairs, it improves small-object Recall@10
from 0.347 to 0.415 ($+19.6\%$, $p<0.01$) and overall Recall@10
from 0.544 to 0.566 ($+4.1\%$, $p<0.05$), with a non-significant
large-object decrease ($-6.0\%$, $p=0.10$; Fig.~\ref{fig:correction}).

Pooling small-object and large-object detections into one ranking
(necessary since within-group ranking is invariant to the
correction), Average Precision decreases from 0.500 to 0.476
($\Delta={-}0.025$, 95\% bootstrap CI $[{-}0.041,{-}0.007]$,
$p=0.002$): promoting small objects surfaces more false positives.
This does not conflict with the recall gain: Recall@$k$ only
checks the top $k$, while AP integrates precision across the full
ranking, including the low-confidence tail where reordering hurts
most.

Calibration also improves (Table~\ref{tab:calib_corr}): ECE
roughly halves for small ($0.102\!\to\!0.047$) and medium
($0.079\!\to\!0.040$) objects, and is unchanged for the large-object
reference. Overall, the correction trades some ranking sharpness
for better-calibrated, more equitable small-object detections,
with no retraining or detector-specific assumptions.

\begin{table}[h]
\vspace{-5mm}
\caption{Calibration before and after the temperature-scaling
correction, on the same detection set as
Fig.~\ref{fig:calibration}.
ECE: Expected Calibration Error.
Large objects are the uncorrected reference group
($T_\mathrm{large}=1$) and are unaffected by construction.}
\label{tab:calib_corr}
\centering
\begin{tabular}{lcccc}
\toprule
Scale & Raw gap & Corrected gap & Raw ECE & Corrected ECE \\
\midrule
Small  & $-0.384$ & $-0.225$ & 0.102 & 0.047 \\
Medium & $-0.281$ & $-0.269$ & 0.079 & 0.040 \\
Large  & $-0.205$ & $-0.205$ & 0.110 & 0.110 \\
\bottomrule
\end{tabular}
\end{table}

\newpage
\section{Discussion and Conclusion}
\label{sec:conclusion}

Foundation models have enabled remarkable open-vocabulary
detection capabilities, but their confidence scores carry a hidden
cost.
The same image-level contrastive objective that makes CLIP
powerful for semantic matching also makes its cosine similarity
scores unreliable as region-level localization probabilities.
Taken in isolation, the direction of each bias is
unsurprising. Large objects score higher, and generic queries
score lower.
What we contribute is the quantification beyond that direction.
First, a consistent, statistically significant scale-bias effect
across three architectures, corroborated by a large-scale
replication on 1{,}203 LVIS categories, alongside a comparably
significant semantic-bias effect on GroundingDINO.
Second, a structural derivation (Sec.~\ref{sec:theory}) showing
both biases arise from the same angular-concentration mechanism,
rather than being independent coincidences.
Third, a demonstration that the resulting loss of discriminative
power is irreversible by threshold selection alone
(Consequence~1).
Together, these results make the entanglement a structural
consequence of adapting image-level representations to
region-level tasks. Threshold tuning alone cannot fix it.

\subsubsection{Scope and Limitations.}
Three limitations qualify these contributions.
First, the ideal confidence score $\sstar$
(Sec.~\ref{sec:formulation}) is defined purely as localization
correctness, $P(\mathrm{IoU}\geq0.5\mid v,t)$.
We choose this definition because it is the quantity a
practitioner implicitly assumes when filtering detections with a
single threshold.
We do not claim semantic-grounding uncertainty is irrelevant.
It is simply a distinct axis from the localization signal our
diagnosis targets, and conflating the two inside one scalar is
itself part of the problem we describe.
Second, a potential confound is that large objects may be easier
to detect due to lower occlusion.
A dedicated experiment across 640 large objects rules this out:
$r=-0.000$, $p=0.99$ under our occlusion proxy (COCO segmentation
area ratio).
Other correlates of scale, namely local resolution, background
clutter, candidate-proposal quality, and category frequency, are
not individually disentangled from the mechanism we identify, and
may contribute alongside it.
Our LVIS frequency-stratified analysis (Fig.~\ref{fig:lvis}, right)
is a first step toward separating category frequency from pure
scale, but a full disentanglement is future work.
Future work should also identify which architectural component,
the spatial pooling stage or the contrastive alignment objective,
is most responsible for each bias.
Third, all controlled experiments use COCO 2014 val, with a
GroundingDINO-based LVIS replication.
Our three-detector cross-architecture replication
(Table~\ref{tab:cross}) and the structural argument in
Sec.~\ref{sec:theory} both support the claim that the same
direction holds for any detector built on image-level
vision-language features.
Broader validation across additional datasets, architectures, and
deployment scenarios remains future work.
These limitations narrow what specific numbers we can claim.
They do not change the structural argument: it is not a sampling
artefact, and its practical consequences follow directly.
\newpage
The entanglement has direct implications for deployment.
Semantic bias means that confidence scores are exquisitely
sensitive to prompt wording (``a dog'' vs.\ ``an animal'').
The same detection can be accepted or rejected depending on how a
user phrases the query.
This prompt sensitivity, a known concern for foundation model
applications, here has a concrete, quantifiable effect on
detection reliability.
Any downstream system in robotics, video surveillance, or medical
imaging that relies on open-vocabulary confidence scores will
systematically underserve small objects and over-represent large
ones.
Concrete examples include distant-pedestrian detection for
autonomous driving, small-instrument tracking in surgical or
industrial inspection footage, and rare, small-object categories in
aerial or satellite imagery.
In these long-tail settings, scale bias and the semantic effect we
observe on LVIS (Fig.~\ref{fig:lvis}) compound.
As the community moves toward lightweight
detectors~\cite{cheng2024yolo}, this problem becomes more acute.
Smaller models cannot afford post-hoc VLM verification, which makes
well-calibrated scores a prerequisite rather than a luxury.

Our temperature scaling correction shows that scale bias is
partially reversible at inference time without architectural
change.
This offers a practical mitigation while the deeper solution
awaits: detectors that predict $P(\mathrm{IoU} \geq 0.5 \mid v, t)$
directly via region-level supervision, rather than inheriting
image-level similarity as a proxy.
As Sec.~\ref{sec:consequences} shows, this correction is not free.
It trades a modest, statistically significant precision cost for
substantially better calibration on the objects it targets.
Extending it to a larger pool of images, across detectors, and on
a downstream task remains future work.

Confidence scores from foundation-model-based open-vocabulary
detectors should not be treated as calibrated localization
probabilities without correction. A confidence score
that depends on how large an object happens to be, or on which of
several equivalent words a user chose, is not measuring confidence:
it is measuring scale and wording, dressed up as certainty.

\bibliographystyle{splncs04}
\bibliography{references}

\end{document}